\documentclass[acmsmall,review=False,anonymous=False,natbib=True,screen=True]{acmart}\usepackage[]{graphicx}\usepackage[]{color}
\makeatletter
\def\maxwidth{ %
  \ifdim\Gin@nat@width>\linewidth
    \linewidth
  \else
    \Gin@nat@width
  \fi
}
\makeatother

\definecolor{fgcolor}{rgb}{0.345, 0.345, 0.345}

\usepackage{framed}
\makeatletter
 {\par\unskip\endMakeFramed%
 \at@end@of@kframe}
\makeatother

\definecolor{shadecolor}{rgb}{.97, .97, .97}
\definecolor{messagecolor}{rgb}{0, 0, 0}
\definecolor{warningcolor}{rgb}{1, 0, 1}
\definecolor{errorcolor}{rgb}{1, 0, 0}
\newenvironment{knitrout}{}{} 

\usepackage{alltt}

\definecolor{c77a1d2}{RGB}{119,161,210}
\definecolor{bf9837}{RGB}{191,152,55}
\definecolor{cc0c0c0}{RGB}{192,192,192}

\usepackage[american]{babel}
\usepackage{tikz}
\usepackage{booktabs}
\usepackage{multicol}
\usepackage{subcaption}
\AtBeginDocument{%
  \providecommand\BibTeX{{%
    \normalfont B\kern-0.5em{\scshape i\kern-0.25em b}\kern-0.8em\TeX}}}
    
\copyrightyear{2021}
\acmYear{2021}
\setcopyright{acmlicensed}\acmConference[OpenSym 2021]{17th International Symposium on Open Collaboration}{September 15--17, 2021}{Online, Spain}
\acmBooktitle{17th International Symposium on Open Collaboration (OpenSym 2021), September 15--17, 2021, Online, Spain}
\acmPrice{15.00}
\acmDOI{10.1145/3479986.3479991}
\acmISBN{978-1-4503-8500-8/21/09}

\usepackage{xcolor}
\usepackage{colortbl}
\definecolor{mygreen}{HTML}{43bf71}
\usepackage{subcaption}

\let\oldciteauthor=\citeauthor
\def\citeauthor#1{{\hypersetup{citecolor=black}\oldciteauthor{#1}}}
\usepackage[htt]{hyphenat}
\usepackage{amsmath}
\IfFileExists{upquote.sty}{\usepackage{upquote}}{}
\begin{document}


\title[Measuring Article Quality]{Measuring Wikipedia Article Quality in One Dimension by Extending ORES with Ordinal Regression}

\author{Nathan TeBlunthuis}
\email{nathante@uw.edu}
\orcid{0000-0002-3333-5013}
\affiliation{%
  \institution{University of Washington}
  \streetaddress{Box 353740}
  \city{Seattle}
  \state{Washington}
  \country{USA}
  \postcode{98195}
}

\renewcommand{\shortauthors}{TeBlunthuis}
    
    
\begin{abstract}
Organizing complex peer production projects and advancing scientific knowledge of open collaboration each depend on the ability to measure quality.  Article quality ratings on English language Wikipedia have been widely used by both Wikipedia community members and academic researchers for purposes like tracking knowledge gaps and studying how political polarization shapes collaboration. Even so, measuring quality presents many methodological challenges. The most widely used systems use labels on discrete ordinal scales when assessing quality, but such labels can be inconvenient for statistics and machine learning. Prior work handles this by assuming that different levels of quality are ``evenly spaced'' from one another. This assumption runs counter to intuitions about the relative degrees of effort needed to raise Wikipedia encyclopedia articles to different quality levels. Furthermore, models from prior work are fit to datasets that oversample high-quality articles. This limits their accuracy for representative samples of articles or revisions. I describe a technique extending the Wikimedia Foundations' ORES article quality model to address these limitations. My method uses weighted ordinal regression models to construct one-dimensional continuous measures of quality. While scores from my technique and from prior approaches are correlated, my approach improves accuracy for research datasets and provides evidence that the ``evenly spaced'' assumption is unfounded in practice on English Wikipedia. I conclude with recommendations for using quality scores in future research and include the full code, data, and models.

\end{abstract}





\maketitle


\section{Introduction} \label{sec:introduction}

Measuring content quality in peer production projects like Wikipedia is important so projects can learn about themselves and track progress. Measuring quality also helps build confidence that information is accurate and supports monitoring how well an encyclopedia includes diverse subject areas to identify gaps needing attention \cite{redi_taxonomy_2021}. Measuring quality enables tracking and evaluating the progress of subprojects and initiatives organized to fill the gaps \cite{halfaker_interpolating_2017, warncke-wang_success_2015}.   Raising an article to a high standard of quality is a recognized achievement among contributors, so assessing quality can help motivate contributions \cite{ayers_how_2008,forte_why_2005}. In these ways, measuring quality can be of key importance to advancing the priorities of the Wikimedia movement and is also important to other kinds of open collaboration \cite{champion_underproduction_2021}.

Measuring quality also presents methodological and ontological challenges.  How can ``quality'' be conceptualized so that measurement of the goals of a project and the value it produces can be precise and accurate?
Language editions of Wikipedia, including English, peer produce quality labels that have been useful both for motivating and coordinating project work and for enabling research.
Epistemic virtues of this approach stem from the community-constructed criteria for assessment and from formalized procedures for third-party evaluation organized by WikiProjects. These systems also have two important limitations: (1) ratings are likely to lag behind changes in article quality, and (2) quality is assessed on a discrete ordinal scale, which violates typical assumptions in statistical analysis. Both limitations are surmountable.

The machine learning framework introduced by \citeauthor{warncke-wang_tell_2013} \cite{warncke-wang_tell_2013}, further developed by \citeauthor{halfaker_interpolating_2017} \cite{halfaker_interpolating_2017}, implemented by the Objective Revision Evaluation Service\footnote{\url{https://www.mediawiki.org/wiki/ORES} (\url{https://perma.cc/TH6L-KFT6})} (ORES) article quality models and adopted by several research studies of Wikipedia article quality \cite[e.g.][]{halfaker_ores_2020, kocielnik_reciprocity_2018, shi_wisdom_2019, warncke-wang_success_2015} was designed to address the first limitation by using article assessments at the time they were made as ``ground truth.'' Article quality might drift in the periods between assessments, but it seems safe to assume that new quality assessments are accurate at the time they are made. A model trained on recent assessments can predict what quality label an article would receive if assessed in its current state.


This paper introduces a method for constructing interpretable one-dimensional measures of article quality from Wikipedia quality assessments and the ORES article quality model. The method improves upon prior approaches in two important ways. First, by using inverse probability weighting to calibrate the model, it is more accurate for typical research applications, and second, it does not depend on the assumption that quality levels are ``evenly spaced,'' which threatens the validity of prior research \cite{halfaker_interpolating_2017, arazy_evolutionary_2019}.  In addition, this paper helps us understand the validity of previous work by analyzing the performance of the ORES quality model and testing the ``evenly spaced'' assumption.

In §\ref{sec:background}, I provide a brief overview of quality measurement in peer production research, in which I foreground the importance of the assumptions needed to use machine learning predictions in downstream analysis---particularly the ``evenly spaced'' assumption used by  \citeauthor{halfaker_interpolating_2017} \cite{halfaker_interpolating_2017} to justify the use of a handpicked weighted sum to combine article class probabilities.  Next, in §\ref{sec:methods}, I describe how to build accurate ordinal quality models that are appropriately calibrated for analyses of representative samples of Wikipedia articles or revisions. I also briefly explain how ordinal regression provides an interpretable one-dimensional measure of quality and how it relaxes the ``evenly spaced'' assumption.  Finally, in §\ref{sec:results} I present the results of my analysis to (1) show how the precision of the measurement depends on proper calibration and (2) demonstrate that the ``evenly spaced'' assumption is violated. Despite this, I find that scores from the ordinal models are highly correlated with those from prior work so the ``evenly spaced'' assumption may be acceptable in some applications. I conclude in §\ref{sec:discussion} with recommendations for measuring article quality in future research.

\section{Background}
\label{sec:background}


Measurement is important to science as available knowledge often constrains the development of improved tools for advancing knowledge.  For example, in the book \textit{Inventing Temperature}, Hasok \citeauthor{chang_inventing_2004} \cite{chang_inventing_2004}, the philosopher and historian of science,  documents how extending theories of heat beyond the range of human sense perception required scientists to develop new types of thermometers. This in turn required better knowledge of heat and of thermometric materials such as the freezing point of mercury.   Part of the challenge of scientific advancement is that measurement devices developed under certain conditions may give unexpected results outside of the range in which they are calibrated: a thermometer will give impossibly low temperature readings when its mercury unexpectedly freezes. Today, machine learning models are used to extend the range of quality measurements in peer production research, but state of the art machine learning can be quite sensitive to the nuances of how their training data are selected \cite{recht_imagenet_2019}. 

\subsection{Measuring Quality in Peer Production}

As described in §\ref{sec:introduction}, measuring quality has been of great importance to peer production projects like Wikipedia and in the construction of knowledge about how such projects work. The foundation of article quality measurement in Wikipedia has been the peer production of article quality assessment organized by WikiProjects who develop criteria for articles in their domain \cite{phoebe_ayers_how_2008}. This enables quality assessment to be consistent across different subject areas, but the procedures for assessing quality are tailored to the values of each WikiProject.  Yet, like human sense perception of temperature, these quality assessments are limited in that they require human time and attention. In addition, humans' limited ability to discriminate between levels on a scale limits the sensitivity of quality assessments. Articles are assessed irregularly and infrequently at the discretion of volunteer editors. Therefore, for most article revisions, it is not known what quality class the article would be assigned if it were newly assessed.

Researchers have proposed many ideas to extend the range of quality measurement beyond the direct perception of Wikipedians, such as page length \cite{blumenstock_size_2008}, persistent word revisions \cite{adler_content-driven_2007, biancani_measuring_2014}, collaboration network structures \cite{raman_classifying_2020}, and template-based flaw detection \cite{anderka_predicting_2012}. Carefully constructed indexes benchmarked against English language Wikipedia quality assessments might allow quality measurement of articles that have not been assessed or in projects that have underproduced article assessments \cite{lewoniewski_relative_2017}. However, such indexes may lack emic validity if they fail to capture important aspects of quality or if notions of quality vary between linguistic communities and might even shape the editing activity in unexpected ways that could ultimately defeat their purpose \cite{goodhart_problems_1984,strathern_improving_1997}. Peer-produced quality labels depend on the limited capacity of volunteer communities to coordinate quality assessment, but also provide impressive validity for evaluating projects on their own terms.  

\subsection{Article Quality Models Extend Measurement to Unassessed Articles} 

Perhaps the most successful approaches to extending the range of quality measurements use machine learning models trained on available article quality assessments to predict the quality of revisions that have not been assessed.  The ORES article quality model (henceforth ORES) implements this approach, but other similar article quality predictors have been developed \cite{anderka_breakdown_2012,dang_quality_2016,zhang_history-based_2018,druck_learning_2008,sarkar_stre_2019,raman_classifying_2020}, and additional features including those based on language models can substantially improve classification performance compared to ORES \cite{schmidt_article_2019}. The ORES model is a tree-based classifier that predicts the quality class of a Wikipedia article at the time it is assessed.\footnote{The system uses cross-validation to select among candidates that include random-forest and boosted decision tree models.} These tree-based models are reasonable for practical purposes with the reported ability to predict within one level of the true quality class with 90\% accuracy (although in §\ref{sec:accuracy} I find a decline in accuracy in a more recent dataset). Yet, since these models do not account for the ordering of quality labels, the use of these predictions in downstream analysis introduces complicated methodological challenges.

The ORES classifiers are fit using \texttt{scikit-learn}\footnote{\url{https://scikit-learn.org/stable/}(\url{https://perma.cc/5Y8B-W8T5})} through minimization of the multinomial deviance as shown \cite{pedregosa_scikit-learn_2011,hastie_elements_2018}:
\begin{equation}
  L(y_i,p(x_i)) = -\sum_{k=1}^K{I(y_i=\mathcal{G}_{i,k})\mathrm{log}~p_k(x_i)}
\label{eq:multinomial.loglik}
\end{equation}

\noindent For each article $i$ with predictors $x_i$ that has been labeled with a quality class $y_i$, the ORES model outputs an estimated probability $p_k(x_i)$ that the article belongs to each quality class $k \in \{\mathrm{\textit{stub}}, \mathrm{\textit{start}}, \mathrm{\textit{C-class}}, \mathrm{\textit{B-class}}, \mathrm{\textit{Good article (GA)}}, \mathrm{\textit{Featured article (FA)}}\}$. The predicted probabilities $p(x_i)$ sum to one so the ORES model outputs a unit vector for each article.  If  $\mathcal{G}_{i,k}$, the most probable quality class (MPQC) according to the model, is the true label, then $I(y_i=\mathcal{G}_{i,k})$ equals $1$ ($I$ is the indicator function) and the log predicted probability $p_k(x_i)$ of the correct class is subtracted from the loss $L(y_i,p(x_i))$. Note that this  model does not use the fact that article quality classes are ordered.  If it did, then it would have to penalize an incorrect classification of a \textit{Good article} as \textit{C-class} more than a classification of a \textit{Good article} as \textit{B-class}. In this model, different quality classes have no intrinsic rank or ordering and thus are akin to different categories of article subjects like animals, vegetables, or minerals.

The MPQC is perhaps the most natural way to use the ORES output to measure quality. It has been used in several studies including to provide evidence that politically polarized collaboration on Wikipedia leads to high quality articles \cite{shi_wisdom_2019} and to understand the relationship between article quality and donation \cite{kocielnik_reciprocity_2018}.  However, the MPQC is limited in that it does not measure quality differences between articles that have the same MPQC. Consider two hypothetical articles; the first has the multinomial prediction $(0.1,0.3,0.4,0.075,0.075,0)$ and the second has the prediction $(0.075,0.075,0.4,0.3,0.1,0)$. The MPQC will assign both the \textit{C-class} label even though the first article has an even chance at being a \textit{Stub} or \textit{Start-class} while the second article has an even chance at being a \textit{B-class} or even a \textit{Good article}.  At best, the MPQC has limited sensitivity to subtle variations or gradual changes in quality \cite{halfaker_interpolating_2017}.  

\subsection{Combining Scores for Granular Measurement}

To further extend the range of article quality measurement within article quality classes, \citeauthor{halfaker_interpolating_2017} \cite{halfaker_interpolating_2017} constructed a numerical quality score using a linear combination (a weighted sum) of the elements of the multinomial prediction $p(x_i)$. This is advantageous from a statistical perspective as it naturally provides a continuous measure of quality which can typically justify a normal or log-normal statistical model. It can also support higher-order aggregations for measuring the quality of a set of articles \cite{halfaker_interpolating_2017}.  \citeauthor{halfaker_interpolating_2017} handpicks the coefficients $[0,1,2,3,4,5]$ to make a linear combination of the predictions under the assumption ``that the ordinal quality scale developed by Wikipedia editors is roughly cardinal and evenly spaced,'' which I refer to the ``evenly spaced'' assumption. It essentially says that a \textit{Start-class} article has one more unit quality of a \textit{Stub-class} article, and that a \textit{C-class} article has one more unit of quality than a \textit{Start-class} article and so on. This approach is being adopted by other researchers including  \citeauthor{arazy_evolutionary_2019} \cite{arazy_evolutionary_2019}.

The considerable degree of effort and expertise required to raise articles to higher levels of quality raises doubt in the assumption \cite{jemielniak_common_2014}.  Higher quality levels correspond to increasing completeness, encyclopedic character, usefulness to wider audiences, incorporation of multimedia, polished citations, and adherence to Wikipedia's policies. The English language Wikipedia editing guideline on content assessment\footnote{\url{https://en.wikipedia.org/w/index.php?title=Wikipedia:Content_assessment&oldid=1023695750} (\url{https://perma.cc/2JUV-6SD})} defines a \textit{Good article} as ``useful to nearly all readers, with no obvious problems'' and a \textit{Featured article} article as ``professional, outstanding and thorough.'' According to Wikipedians, it can take ``three to six months of full time work'' to write a \emph{Featured article}.\footnote{Public statement by Stuart Yeates, an expert Wikipedian; quoted with permission. \url{https://lists.wikimedia.org/hyperkitty/list/wiki-research-l@lists.wikimedia.org/message/7U35LHAXRWEPABN75DOTPOIEA2VYCTQQ/} (\url{https://perma.cc/9V4P-WRXR})}  Are we to assume that the difference in quality between a \textit{Good article} and a \textit{Featured article} is measurably the same as that between a \textit{Stub} defined as as ``little more than a dictionary definition'' and a \textit{Start-class} that is ``a very basic description of the topic?'' How could we even answer this question? 

If the ``evenly spaced'' assumption is reasonable, then \citeauthor{halfaker_interpolating_2017}'s \cite{halfaker_interpolating_2017} weighted sum approach is too. But if increasing Wikipedia article classes do not represent roughly equal improvements in quality, this may threaten the accuracy of analysis dependent on the assumption. Suppose that a \textit{B-class} has not 1, but 2 units of quality greater than a \textit{C-class} article, then \citeauthor{halfaker_interpolating_2017} could have underestimated the improvement in the knowledge gap of women scientists, which was considerably driven by improvement in \textit{B-class} articles.  In the next section, I provide a straightforward extension of the ORES article quality model based on ordinal regression can both relax the ``evenly spaced'' assumption and provide a better calibrated and more accurate one-dimensional measure of quality.


\section{Data, Methods and Measures}
\label{sec:methods}


I use Bayesian ordinal regression models that use the ORES predicted probabilities to predict the quality class labels and quantify the distance between quality classes. I now provide a brief overview of ordinal regression as needed to explain my approach to measuring quality. Understanding ordinal regression depends on background knowledge of odds and generalized linear models.  I recommend \citeauthor{mcelreath_statistical_2018} \cite{mcelreath_statistical_2018} for reference.

\subsection{Bayesian Ordinal Regression}

Ordinal regression predicts quality class membership using a single linear model for all classes and identifies boundaries between classes using the log cumulative odds link function shown below in Eq. \ref{eq:ordinal.regression}.  The log cumulative odds is not the only possible choice of link function, but it is the most common, is the easiest to interpret, and is appropriate here.

\begin{align}
  \mathrm{log}&~\frac{\mathrm{Pr}(y_i \le k)}{1 - \mathrm{Pr}(y_i \le k)} = \alpha_k - \phi_i  \label{eq:ordinal.regression} \\
  \phi_i &= B x_i \nonumber 
\end{align}
\noindent As in Eq. \ref{eq:multinomial.loglik}, $y_i$ is the quality label for article $i$. The left hand side of Eq. \ref{eq:ordinal.regression} gives the log odds that $y_i$ is less than or equal to quality level $k$. The ordinal quality measure is given by a linear model $\phi_i = B x_i$ ($x_i$ is a vector of transformed ORES scores for article $i$).  Key to interpreting $\phi_i$ as a quality measure are the intercept parameters $a_k$ for each quality level $k$. The log cumulative odds (the log odds that the article $y_i$ has quality less than or equal to $k$) are given by the difference between the intercept and the linear model $a_k$ - $\phi_i$. Therefore, if $\phi_i = \alpha_k$ then the chances that $i <= k$ equal the chances that $i > k$. When $\phi_i$ is less than $\alpha_k$, the quality of article $i$ is probably less than or equal to quality level $k$.  As $\phi_i - \alpha_k$ increases so do the chances that article $i$ is of quality better than $k$. In this way, the threshold parameters $a_k$ define quantitative article quality levels on the scale of the ordinal quality measure $\phi_i$.

Informally, an ordinal regression model maps a linear regression model to the ordinal scale using the log cumulative odds link function. It does this by inferring thresholds that partition the range of linear predictions. When the linear predictor for an article crosses a threshold, the probability that the article has quality greater than that corresponding to the threshold begins to increase.

Bayesian inference allows interpreting model parameters like $\phi_i$ and $\alpha_k$ as random variables and provides accurate quantification of uncertainty in thresholds and predictions. I fit models using the R package Bayesian regression modeling using Stan (\texttt{brms}) \cite{burkner_brms_2017} version 2.15.0. I use the default priors for ordinal regression, which are weakly informative.  Due to the large sample size, the data overwhelm the priors and the priors have little influence over results.  I confirmed this by fitting equivalent frequentist models using the \texttt{polr} function in the \texttt{MASS} R package \cite{venables_modern_2002} and found that the estimates of intercepts and coefficients were very close. 

The six quality scores output by the ORES article quality classifier are perfectly collinear by construction because they sum to one. This means they cannot all be included in the same regression model.  Since interpreting the coefficients is not important, I take the linear transformation of the ORES scores using appropriately weighted principle component analysis and use the first five principle components as the independent variables. This is simpler and more statistically efficient than a model selection procedure.


\subsection{Dataset and Model Calibration}
\label{sec:data}

I draw a new random sample of 5,000 articles from each quality class to develop my models. I first reuse code from the \texttt{articlequality}\footnote{\url{https://pypi.org/project/articlequality} (\url{https://perma.cc/8R4H-MAZ9})} Python package to process the March 2020 XML dumps for English Wikipedia and extract up-to-date article quality labels.  I then select pages that have been assessed by a member of at least one WikiProject. Following prior work, if an article is assessed at different levels according to more than one WikiProject, I assign it to the highest such level and I drop articles having the rarely used \emph{A-class} quality level \cite{halfaker_interpolating_2017,warncke-wang_success_2015,warncke-wang_tell_2013}.  Next, I use the \texttt{revscoring}\footnote{\url{https://pypi.org/project/revscoring} (\url{https://perma.cc/3HFN-V23Z})} Python package to obtain the ORES scores of the labeled article versions. Some of these versions have been deleted leading to missing observations at each quality level. Table \ref{tab:sample} shows the number of articles sampled in each quality class. I reserve a random sample of 2000 articles which I use in reporting my results and fit my ordinal regression models on the remainder.   

The ORES article quality classifiers are fit on a ``balanced'' dataset having an equal number of articles in each quality class.  Thus, an ORES score is the probability that an article is a member of a quality class under the assumption that the article was drawn from a population where each quality class contains an equal number of articles.  Simply put, the model has learned from its training data that each quality class is about the same size. 

\begin{figure}
\begin{knitrout}
\definecolor{shadecolor}{rgb}{0.969, 0.969, 0.969}\color{fgcolor}
\includegraphics[width=\maxwidth]{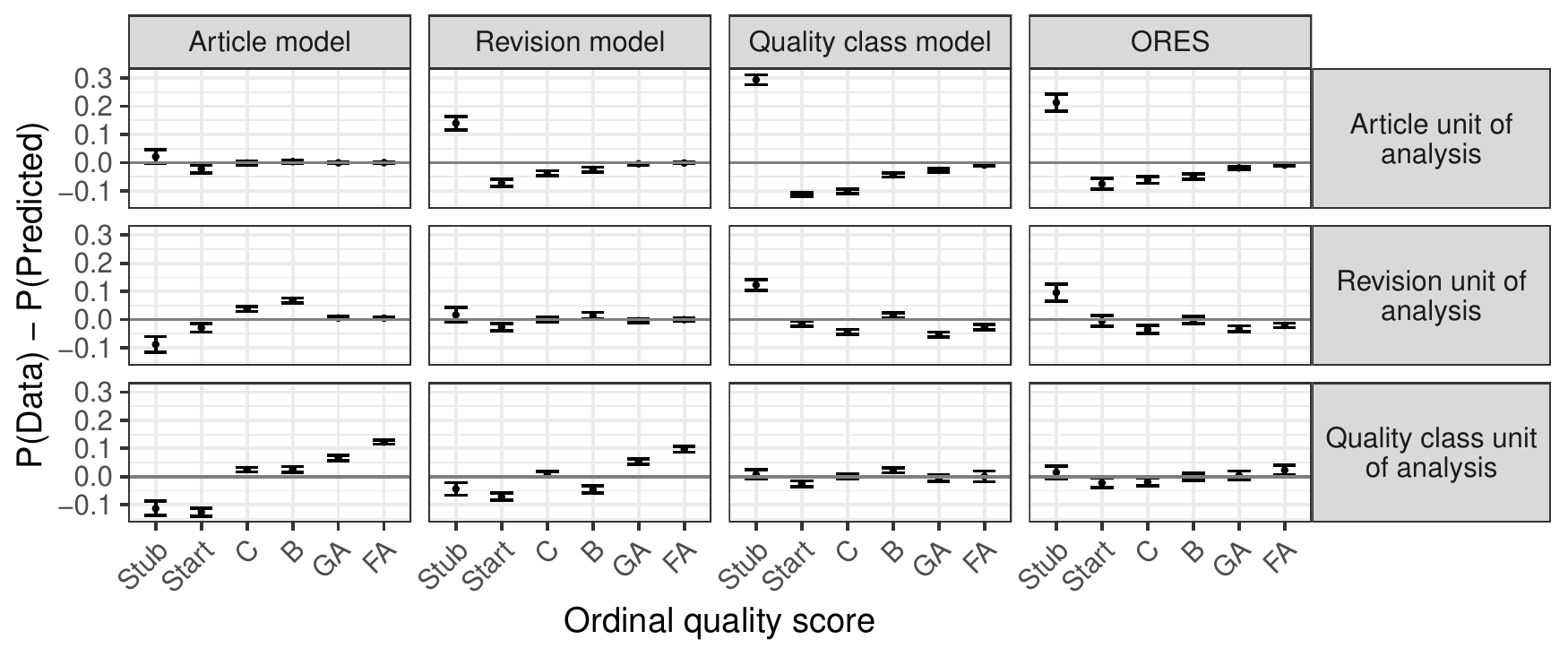} 
\end{knitrout}
\caption{Calibration of each predictive quality model on datasets representative of each unit of analysis (article, revision, quality class). Each chart shows, for each quality class, the miscalibration of a model (columns) with respect to a dataset weighted to represent a unit of analysis (rows). The y-axis shows difference between the true probability of the quality class and the average predicted probability of that class, given a chosen unit of analysis. Points close to zero indicate good calibration. For example, the top-left chart shows that the article model is well-calibrated to the dataset on which it was fit and the middle-left chart shows that the article model predicts that articles are \textit{Stubs} with probability greater than the frequency of \textit{Stubs} in a random sample of revisions. Error bars show 95\% confidence intervals. \label{fig:calibration}}
\end{figure}

This is not representative of the overall article quality on Wikipedia, which is highly skewed with over 3 million \textit{Stubs} but only around \textit{7,000} \textit{Featured articles} as shown in Table \ref{tab:sample}. Although using a balanced dataset likely improves the accuracy of the ORES models, for the ordinal regression models, the choice of unit of analysis presents a trade-off between accuracy in a representative sample of articles or revisions and accuracy within each quality class. 
Constructing a balanced dataset by oversampling is a common practice in machine learning because it can improve predictive performance. However, oversampling can also lead to badly calibrated predictive probabilities as shown in Fig. \ref{fig:calibration}.  Calibration means that, on average, the predicted probability of a quality class equals the average true probability of that class for the unit of analysis. 

The ``balanced'' dataset on which ORES is trained has the \textit{quality class} unit of analysis because each quality class has equal representation. However, researchers are more interested in analyzing representative samples of \textit{articles} or \textit{revisions}. For example, the article unit of analysis would be used to estimate the average quality of a random sample of articles and the revision unit of analysis might be used to model the change in the quality of an encyclopedia over time. 
Weighting allows the use of the balanced dataset to estimate a model as if the dataset were a uniform random sample of a different unit of analysis.
My method uses a balanced dataset to fit ordinal regression models with inverse probability weighting to calibrate each model to the unit of analysis of a research project.
For example, each article in the model calibrated to the article unit of analysis is weighted by the probability of its quality class in the population of articles divided by the probability of its quality class in the sample. The size of the sample and the weights for the article and revision levels of analysis are also shown in Table \ref{tab:sample}.

\begin{table}
\caption{Number of articles sampled at each quality level}
%
\begin{tabular}{lrrrrr}
  \hline
Label & No. of articles & No. of revisions & Sample size & Article weights & Revision weights \\ 
  \hline
Stub & 3,359,351 & 12,005,611 & 4,969 & 4.23 & 2.52 \\ 
  Start & 1,019,038 & 7,828,335 & 4,979 & 1.28 & 1.64 \\ 
  C & 235,655 & 3,889,639 & 4,988 & 0.30 & 0.81 \\ 
  B & 128,875 & 3,640,591 & 4,990 & 0.16 & 0.76 \\ 
  GA & 31,808 & 924,468 & 4,999 & 0.04 & 0.19 \\ 
  FA & 7,438 & 365,255 & 4,995 & 0.01 & 0.08 \\ 
   \hline
\end{tabular}

\label{tab:sample}
\end{table}


\section{Results}
\label{sec:results}
I first report my findings about the spacing of the quality classes in each of the models in §\ref{sec:spacing}.   Quality classes are not evenly spaced, especially when articles or revisions are the unit of analysis.  Next, in §\ref{sec:accuracy}, I report the accuracy of each of the models and the uncertainty of the ordinal quality scale.  All models perform similarly to or better than the MPQC within the pertinent unit of analysis. The unweighted model provides the best accuracy and lowest uncertainty across the entire range of quality levels, but is poorly calibrated for other units of analysis. Finally, in §\ref{sec:correlation}, I show that all quality measures are highly correlated, but the ordinal quality measures agree with one another more than with the ``evenly spaced'' measure.

\begin{figure}
\centering
\begin{knitrout}
\definecolor{shadecolor}{rgb}{0.969, 0.969, 0.969}\color{fgcolor}
\includegraphics[width=\maxwidth]{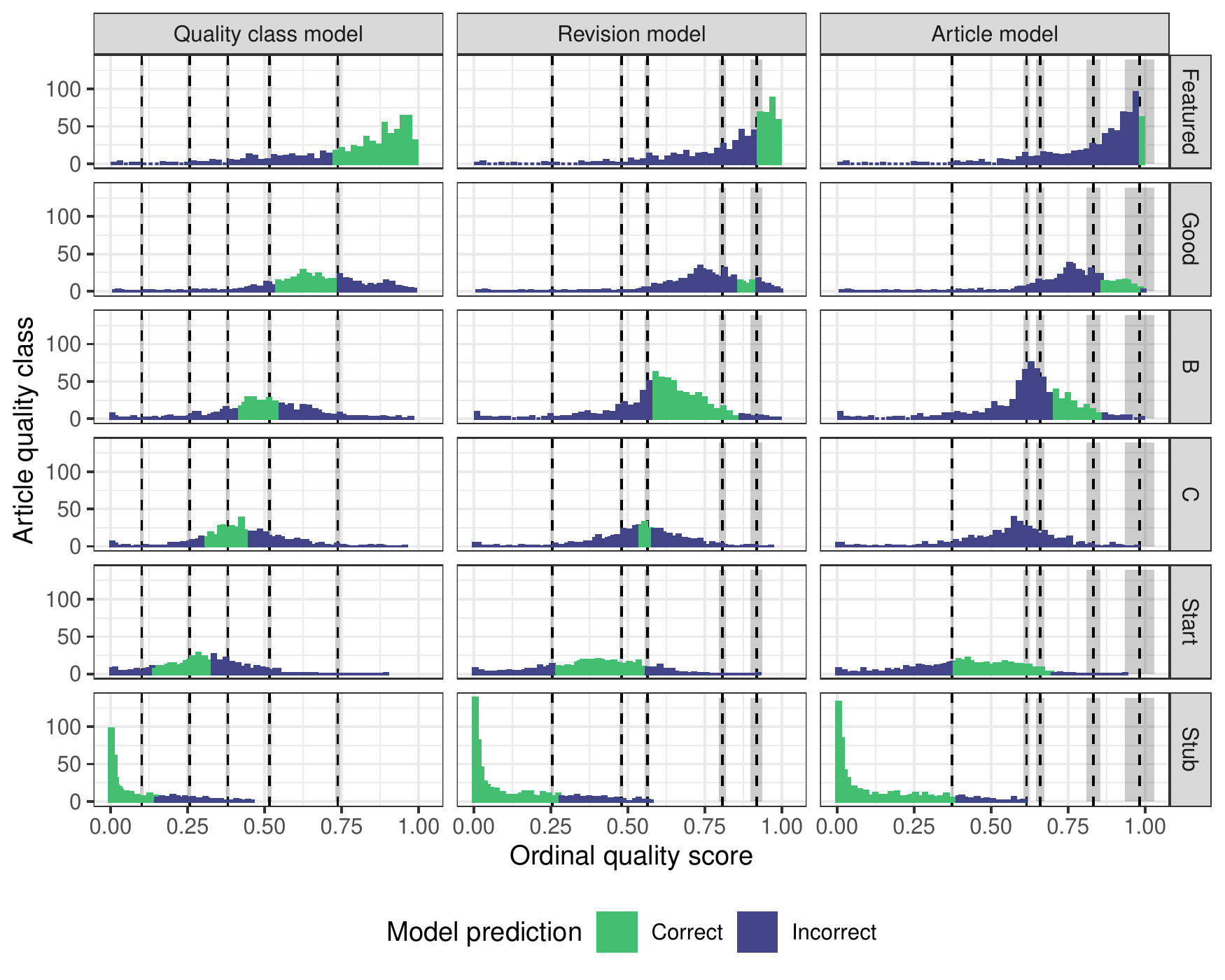} 
\end{knitrout}
\caption{Quality scores and predictions of the ordinal regression models. Columns in the grid of charts correspond to the ordinal quality model calibrated to the indicated unit of analysis and rows correspond to sampled articles having the indicated level of quality as assessed by Wikipedians. Each chart shows the histogram of scores, thresholds inferred by the ordinal model with 95\% credible intervals colored in gray, and colors indicating when the model makes correct or incorrect predictions.  The thresholds are not evenly spaced, especially in \textit{revision model} and \textit{article model} that has more weight on lower quality classes. These two models infer that the gaps between \textit{Stub} and \textit{Start} and between \textit{Start} and \textit{C-class} articles are considerably wider than the gap between \textit{C-class} and \textit{B-class} articles. \label{fig:spacing}}
\end{figure}

\subsection{Spacing of Quality Classes}
\label{sec:spacing}

The grid of charts in Fig. \ref{fig:spacing} shows quality scores and thresholds for each model (columns) and article quality level (rows). Each chart shows the histogram of quality scores $\phi_i$ given to articles having the true quality label corresponding to the row of the grid. The histograms are colored to indicate regions where the model correctly predicts that articles belong to their true class. Vertical dashed lines show the thresholds inferred by the model with 95\% credible intervals colored in gray. Different models have different ranges of scores, so Fig. \ref{fig:spacing} shows results normalized between 0 and 1.

No matter the unit of analysis, article quality classes are not evenly spaced.  The quality class model provides a quality scale in which  \textit{Featured} articles take up $27\%$ of the scale and are expected to score in the range of $[0.73, 1]$, but probable \textit{C-class} articles only span $14\%$ of the scale in the range $[0.31, 0.45]$.  Researchers are likely to be interested in models calibrated to the article or revision units of analysis, and in these cases, the quality classes are far from evenly spaced. The \textit{revision model} assigns $28\%$ of the scale to \textit{Stubs}, from $0$ to $0.28$. It assigns \textit{C-class} articles the smallest part of the scale, only $4\%$ of it, from $0.54$ to $0.58$.  The \textit{article model} is even more extreme. It assigns \textit{Stubs} to the interval $[0, 0.39]$, $39\%$ of the scale, and the space between thresholds defining the range of \textit{C-class} articles is so narrow that it virtually never predicts that an article will be C-class.  In general terms, the \textit{quality class model} gives relatively equal amounts of space to each quality class compared to the other models, while reserving nearly the top half of the scale for the top 2 quality classes.  The \textit{revision model} and \textit{article model} do the opposite and use the bottom half of the scale to account for differences within the bottom two quality classes, leave some room for \textit{B-class} articles, but squeeze the top end of the scale and \textit{C-class} articles into relatively small intervals.


\subsection{Accuracy and Uncertainty}
\label{sec:accuracy}

I evaluate predictive performance in terms of \textit{accuracy}, the proportion of predictions of article quality that are correct.  To allow comparison with the reported accuracy of the ORES quality models, I also report \textit{off-by-one accuracy}, which includes predictions within one level of the true quality class among correct predictions.

\begin{table}
\caption{Accuracy of quality prediction models depends on the unit of analysis. The greatest accuracy and off-by-one accuracy scores are highlighted.  Models are more accurate when calibrated on the same unit of analysis on which they are evaluated.  Compared to the MPQC, the ordinal quality models have better accuracy when revisions or articles are the unit of analysis.  When the quality class is the unit of analysis, the ordinal quality model has worse accuracy, but predicts within one quality class with slightly better accuracy. \label{tab:accuracy}}  
%
\begin{tabular}{lllll}
  \hline
Unit of analysis & Model & Ordinal model? & Accuracy & Off-by-one accuracy \\ 
  \hline
Quality class & Article & Yes & 0.33 & 0.75 \\ 
  Quality class & Revision & Yes & 0.44 & 0.84 \\ 
  Quality class & Quality class & Yes & 0.52 & \cellcolor{mygreen}0.87 \\ 
  Quality class & ORES MPQC & No & \cellcolor{mygreen}0.55 & 0.86 \\ 
   \hline
Revision & Article & Yes & 0.57 & 0.87 \\ 
  Revision & Revision & Yes & \cellcolor{mygreen}0.61 & \cellcolor{mygreen}0.92 \\ 
  Revision & Quality class & Yes & 0.54 & 0.88 \\ 
  Revision & ORES MPQC & No & 0.58 & 0.9 \\ 
   \hline
Article & Article & Yes & \cellcolor{mygreen}0.76 & \cellcolor{mygreen}0.97 \\ 
  Article & Revision & Yes & 0.73 & 0.96 \\ 
  Article & Quality class & Yes & 0.63 & 0.92 \\ 
  Article & ORES MPQC & No & 0.65 & 0.94 \\ 
   \hline
\end{tabular}

 \end{table}

As shown in Table \ref{tab:accuracy}, the ordinal regression models have better predictive ability than the MPQC except when the unit of analysis is the quality class.  In this case, the best ordinal quality model has worse accuracy than the MPQC but slightly better off-by-one accuracy. Table \ref{tab:accuracy} shows accuracy and off-by-one accuracy weighted for each unit of analysis. Accuracy for a given unit of analysis depends on having a model fit to data representative of that unit of analysis. Accuracy scores are higher when greater weight is placed on lower article quality classes, suggesting that it is easier to discriminate between these classes.

The ORES article quality model has been quickly adopted by researchers, but its accuracy is limited.  While off-by-one accuracy is above 90\% when the article is the unit of analysis, the MPQC only predicts the correct quality class 55\% of the time when the quality class is the unit of analysis.     

\begin{figure}
\begin{knitrout}
\definecolor{shadecolor}{rgb}{0.969, 0.969, 0.969}\color{fgcolor}
\includegraphics[width=\maxwidth]{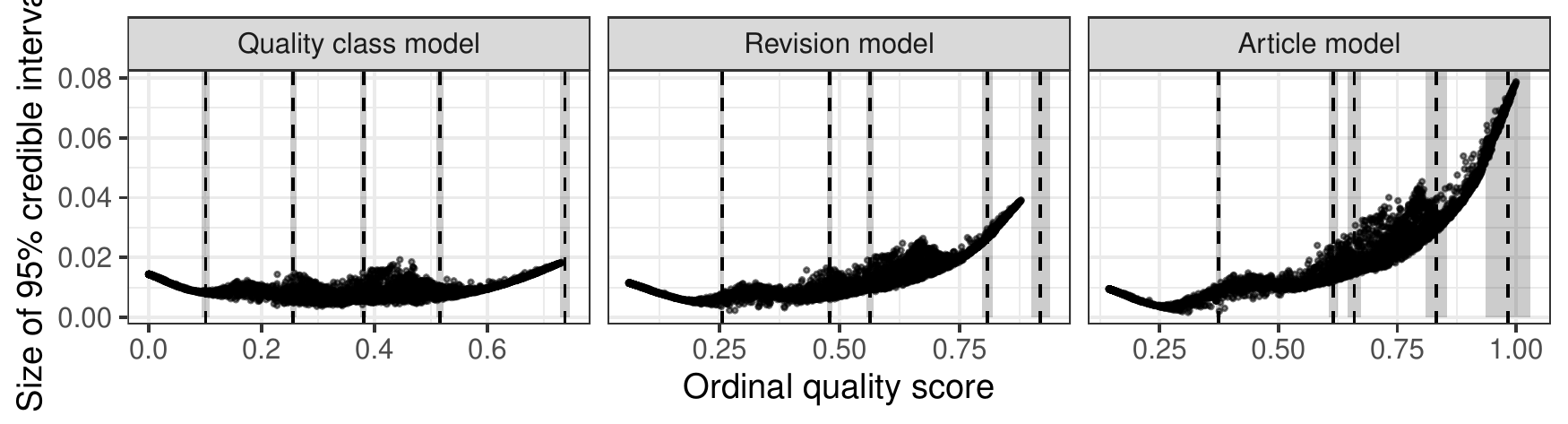} 
\end{knitrout}
\caption{Uncertainty in ordinal quality scores for models calibrated at each unit of analysis.  Points show the size of the 95\% credible interval for the ordinal quality score for each article in the dataset. The quality class model has low uncertainty across the range of quality. Models calibrated to the revision and article levels of analysis have less uncertainty at the low end of the quality scale, but greater uncertainty at the higher end of the scale. \label{fig:uncertainty}}
\end{figure}

The trade-offs in selecting a unit of analysis on which to calibrate the models are further illustrated by Fig. \ref{fig:uncertainty}, which plots the size of the 95\% credible intervals as a function of the quality scores for each model. As in Fig. \ref{fig:spacing}, quality scores in this plot are rescaled between 0 and 1. The models calibrated to articles or revisions have more certainty in the lower range of the quality scale compared to the model that places equal weight in all quality classes.  This comes with a trade-off for the higher range of quality.  While the \textit{quality class model} has relatively low uncertainty across the entire range of quality, the \textit{revision model} and \textit{article model} have greater uncertainty at higher levels of quality.   

\subsection{Correlation Between Scores}
\label{sec:correlation}

\begin{figure}
\begin{knitrout}
\definecolor{shadecolor}{rgb}{0.969, 0.969, 0.969}\color{fgcolor}
\includegraphics[width=\maxwidth]{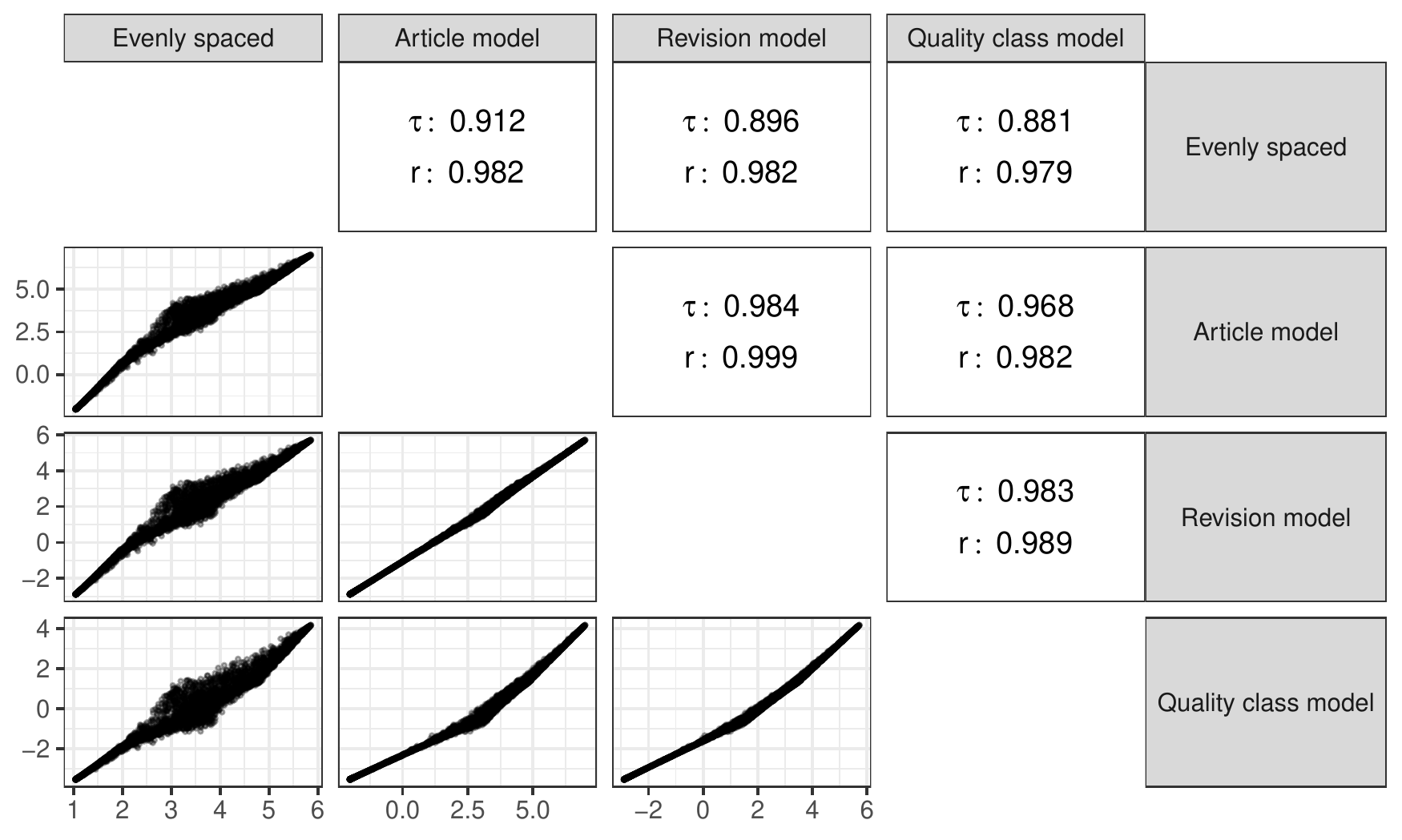} 
\end{knitrout}
  \caption{Correlations between quality measures show that the different approaches to measuring quality are quite similar.  ``Evenly spaced'' uses a weighted sum of the ORES scores with handpicked coefficients \cite{halfaker_interpolating_2017}. Lower values of Kendall's $\tau$, a nonparametric rank correlation statistic, compared to Pearson's $r$ suggest nonlinear differences between the weighted sum and the other measures. \label{fig:correlation}}
\end{figure}

Although the models have different predictive performances and uncertainties, as measures of quality, they are nearly perfectly correlated with one another as shown in Fig. \ref{fig:correlation}. For each quality score, including the ``evenly spaced'' weighted sum, Fig. \ref{fig:correlation} shows a scatter plot and two correlation statistics: Kendall's $\tau$ and Pearson's $r$.  Pearson's $r$ is the standard linear correlation coefficient and Kendall's $\tau$ is a nonparametric rank-based correlation defined as the probability that the quality scores will agree about which of any two articles has higher quality minus the probability that they will disagree.

According to Pearson's $r$ all the quality scores are highly correlated with correlation coefficients of about $0.98$ or higher. Kendall's $\tau$ measures nonlinear correlation and reveals discrepancies between the ordinal models and the ``evenly spaced'' measures. The Pearson correlation between scores from the \textit{revision model} and the scores from the \textit{quality class model} are about the same as the correlation between the \textit{revision model} scores and the  ``evenly spaced'' scores ($r=0.98$). However, according to Kendall's $\tau$, scores from the \textit{revision model} are more similar to those from the \textit{quality class model}  ($r=0.98$) than to the scores from the ``evenly spaced'' approach ($r=0.9$).

The evenly spaced model is more likely to disagree with the model-based scores than any of the model-based scores are to disagree with one another as visualized in the scatter plots in Fig. \ref{fig:correlation}. Disagreement between the ``evenly spaced'' method and the ordinal models is greatest among articles in the middle of the quality range. 

\section{Discussion}
\label{sec:discussion}
Past efforts to extend the measurement of Wikipedia article quality from peer-produced article quality assessments to unassessed versions of articles and from the discrete to the continuous domain have relied upon machine learning and expedient but untested assumptions like that quality levels are ``evenly spaced.''  
While I suggest technical improvements for statistical models for measuring quality, I also find that scores from my models are highly correlated to those obtained under the ``evenly spaced'' assumption.

I set out to provide a better way to convert the probability vector output by the ORES article quality model into a continuous scale and to test the assumption that the quality levels are evenly spaced.  I used ordinal regression models to infer spacing between quality levels and used the linear predictor of these models as a continuous measure of quality.  While I found in §\ref{sec:spacing} that the quality levels are not evenly spaced and that the spacing depends on the unit of analysis to which the models are calibrated, I also showed in §\ref{sec:correlation} that the model-based quality measures are highly, although not perfectly, correlated with  the ``evenly spaced'' measure.  This provides some assurance that past results built on this measure are unlikely to mislead. That said, I recommend that future work adopt appropriately calibrated model-based quality measures instead of the ``evenly spaced'' approach, and I argue that it is important to improve the accuracy of article quality predictors to enable more precise article quality measurement.

\subsection{Recommendations for Measuring Article Quality}
How should future researchers approach the question of how to measure Wikipedia article quality?  While I cannot provide a final or complete answer to the question, I believe the exercise reported in this paper provides some insights on which to base recommendations. It is important to note that I consider here only approaches to measuring quality that assume the use of a good predictor of article quality assessment, such as the ORES quality model.  I do not consider other based approaches such as those based on indexes  \cite{lewoniewski_relative_2017} described in §\ref{sec:background}. 

\subsubsection{Use the principle components of ORES scores for statistical control of article quality}
In many statistical analyses, the only purpose of measuring quality will be as a statistical control or adjustment. For example, \citeauthor{zhang_crowd_2017} \cite{zhang_crowd_2017} used the MPQC as a control variable in a propensity score matching analysis of promotion to \textit{Featured article} status, but as argued in §\ref{sec:methods}, the MPQC provides less information than the vector of ORES scores. Using the principle components is simpler than using an ordinal quality model. I recommend obtaining ORES scores for your dataset, taking the principle components, and dropping the least significant one to remove collinearity. 

\subsubsection{Use ordinal quality scores when article quality is an independent variable}
\label{sec:qciv}
In other cases, research questions will ask how article quality is related to an outcome of interest, like how \citeauthor{kocielnik_reciprocity_2018} \cite{kocielnik_reciprocity_2018} set out to explore factors associated with donations to the Wikimedia Foundation. They use the MPQC as an independent variable, which complicates their analysis.  Although they conclude that ``pages with higher quality attract more donations,'' this is not strictly true. They actually found a nonlinear relationship where readers of \textit{B-class} articles were more likely to donate than readers of \textit{Featured articles}. Using a continuous measure of quality is more convenient when the average linear relationship is the target of inference.  

I recommend using an ordinal regression model appropriate to the downstream unit of analysis because this will justify the interpretation of the measure.  If the downstream unit of analysis differs substantively from those used here, such as if different selection criteria are applied, I recommend reusing my code to calibrate a new ordinal regression model to a new dataset.  Otherwise, reusing one of my models should be adequate. Finally, in the Bayesian framework, the scores are interpretable as random variables.  This provides a justification for incorporating the variance of these scores as measurement errors to improve estimation in downstream analysis \cite{mcelreath_statistical_2018}. 



\subsubsection{Use the MPQC or ordinal quality scores when article quality is the dependent variable}

Using the MPQC as the outcome in an ordinal regression model, as is done by \citeauthor{shi_wisdom_2019} \cite{shi_wisdom_2019} in their analysis of Wikipedia articles with politically polarized editors, is a reasonable choice as long as it provides sufficient variation and a more granular quality measure is not needed. Although it is theoretically possible that using the MPQC might introduce statistical bias because it less accurate than ordinal quality scores for units of analysis other than the quality class and omits variation within quality classes, such threats to validity do not seem more significant than the threat introduced by inaccurate predictions. If the MPQC does not provide sufficient granularity and a continuous measure is desired as in \citeauthor{halfaker_interpolating_2017} \cite{halfaker_interpolating_2017} or \citeauthor{arazy_evolutionary_2019} \cite{arazy_evolutionary_2019}, I recommend using a measure based on ordinal regression as described in §\ref{sec:qciv}.

\subsection{Limitations}

Although intuitions about the varying degrees of effort required to develop articles with different levels of quality led me to question the ``evenly spaced'' assumption, my findings that quality classes are not evenly spaced do not necessarily reflect relative degrees of effort.  Rather, spaces between levels are chosen to link a linear model to ordinal data.  The spacing of intervals depends on the ability of the ORES scores to predict quality classes. The ORES article quality model has relative difficulty classifying \textit{C-class} and \textit{B-class} articles \cite{halfaker_interpolating_2017}. Perhaps, the differences between these quality classes are minor compared to the other classes. Maybe ORES lacks the features or ability to model these differences and the space between these classes will grow if its predictive performance improves.  

The usefulness of article quality scores depends on the accuracy of the model. The ORES quality models are accurate enough to be useful for researchers, but they still only predict the correct quality class 55\% of the time on a balanced dataset.  Of course, this limits the accuracy of the ordinal regression models reported here. 
Furthermore, while the ORES quality models were designed with carefully chosen features intended to limit biases \cite{halfaker_ores_2020}, it is still quite plausible that the accuracy of predictive quality models may vary depending on characteristics of the article \cite{kleinberg_inherent_2016}.  Such inaccuracies may introduce bias, threaten downstream analysis or lead to unanticipated consequences of collaboration tools built upon the models \cite{teblunthuis_effects_2021}. Therefore, improving the accuracy of article quality prediction models is important to the validity of future article quality research. Adopting machine learning models that can incorporate ordinal loss functions is a promising direction and can reduce the need for auxiliary ordinal regression models \cite{cardoso_learning_2007}.

This paper only considers measuring article quality for English language Wikipedia, but expanding knowledge of collaborative encyclopedia production depends on studying other languages as audiences and collaborative dynamics can greatly vary between projects \cite{hecht_tower_2010,lemmerich_why_2019,teblunthuis_dwelling_2019}.  Other languages carry out quality assessments \cite{lewoniewski_relative_2017}, and some of these have been used to build ORES article quality models.  Future work should extend this project to provide multilingual article quality measures in one continuous dimension.

An additional limitation stems from the likelihood that peer-produced quality labels are biased. For instance, the English Wikipedia community has a well-documented pattern of discrimination against content associated with marginalized groups such as biographies of women \cite{tripodi_ms_2021, menking_people_2019} and indigenous knowledge \cite{van_der_velden_decentering_2013}. Although demonstrating biases in article quality assessment is a task for future research, if Wikipedians' assessments of article quality are biased then model predictions of quality will almost certainly be as well.

\section{Conclusion}
Measuring article quality in one continuous dimension is a valuable tool for studying the peer production of information goods because it provides granularity and is amenable to statistical analysis.  Prior approaches extended ORES article quality prediction into a continuous measure under the ``evenly spaced'' assumption. I showed how to use ordinal regression models to transform the ORES predictions into a continuous measure of quality that is interpretable as a probability distribution over article quality levels, provides an account of its own uncertainty and does not assume that quality levels are ``evenly spaced.''  Calibrating the models to the chosen unit of analysis improves accuracy for research applications. I recommend that future work adopt this approach when article quality is an independent variable in a statistical analysis.

\section{Code and Data Availability}
Code, data and instructions for replicating or reusing this analysis are available in the Harvard Dataverse at \url{https://doi.org/10.7910/DVN/U5V0G1}.

\begin{acks}

I am grateful to the members of the Community Data Science Collective for their feedback on early drafts of this work including Kaylea Champion, Sneha Narayan, Jeremy Foote, and Benjamin Mako Hill. I would also like to thank Aaron Halfaker for encouraging me to write this after seeing a preliminary version. Thanks to Stuart Yeates and other participants in the \texttt{wiki-research-l} mailing list \url{wiki-research-l@lists.wikimedia.org} for answering my questions about measuring article quality and effort. Finally, thank you to the anonymous OpenSym reviewers whose careful and constructive feedback improved the paper.

\end{acks}

\bibliographystyle{ACM-Reference-Format}
\bibliography{refs}
\end{document}